\documentclass[10pt,conference]{IEEEtran}

\usepackage{cite}

\ifCLASSINFOpdf
   \usepackage[pdftex]{graphicx}
   \graphicspath{{figs/}}
   \DeclareGraphicsExtensions{.pdf,.jpeg,.png}
\else
   \usepackage[dvips]{graphicx}
   \graphicspath{{../figs/}}
   \DeclareGraphicsExtensions{.eps}
\fi

\usepackage[cmex10]{amsmath}
\usepackage{amsthm}
\usepackage{amsmath}

\usepackage[dvipsnames]{xcolor}
\usepackage{amsfonts}
\usepackage{tabularx}
\usepackage{multirow}
\usepackage{comment}
\usepackage{stfloats}
\usepackage{booktabs}
\usepackage[switch]{lineno}
\usepackage{sistyle}

\usepackage{graphicx}
\usepackage{printlen}

\SIthousandsep{,}

\usepackage{algorithmic}

\usepackage{array}

\ifCLASSOPTIONcompsoc
  \usepackage[caption=false,font=normalsize,labelfont=sf,textfont=sf]{subfig}
\else
  \usepackage[caption=false,font=footnotesize]{subfig}
\fi

\usepackage{url}

\newcommand\ADDED[1]{{\textcolor{black}{#1}}}

\newif\iffinal
\finaltrue
\newcommand{\cmtid}{46}

\iffinal
\else
\usepackage[switch]{lineno}
\fi

\begin{document}

\title{TCDiff: Triple Condition Diffusion Model with\\ 3D Constraints for Stylizing Synthetic Faces}

\iffinal
    \author{
    \IEEEauthorblockN{
    Bernardo Biesseck\IEEEauthorrefmark{1}\IEEEauthorrefmark{2},
    Pedro Vidal\IEEEauthorrefmark{1},
    Luiz Coelho\IEEEauthorrefmark{3},
    Roger Granada\IEEEauthorrefmark{3},
    David Menotti\IEEEauthorrefmark{1}
    }
    
    \IEEEauthorblockA{\IEEEauthorrefmark{1}Depart. of Informatics, Federal University of Paraná, Curitiba, PR, Brazil
    \texttt{\footnotesize \{bjgbiesseck, pbqv20, menotti\}@inf.ufpr.br}}
    
    \IEEEauthorblockA{\IEEEauthorrefmark{2}Federal Institute of Mato Grosso (IFMT), Pontes e Lacerda, Brazil \texttt{\footnotesize \{bernardo.biesseck\}@ifmt.edu.br}}
    
    \IEEEauthorblockA{\IEEEauthorrefmark{3}unico - idTech, Brazil
    \texttt{\footnotesize \{luiz.coelho, roger.granada\}@unico.io}}
    }
\else
    \author{SIBGRAPI Paper ID: \cmtid \\ }
    \linenumbers
\fi

\maketitle

\let\thefootnote\relax\footnote{\\979-8-3503-7603-6/24/\$31.00~\copyright~2024 IEEE\hfill}

\begin{abstract}
A robust face recognition model must be trained using datasets that include a large number of subjects and numerous samples per subject under varying conditions (such as pose, expression, age, noise, and occlusion).
Due to ethical and privacy concerns, large-scale real face datasets have been discontinued, such as MS1MV3, and synthetic face generators have been proposed, utilizing GANs and Diffusion Models, such as SYNFace, SFace, DigiFace-1M, IDiff-Face, DCFace, and GANDiffFace, aiming to supply this demand. 
Some of these methods can produce high-fidelity realistic faces, but with low intra-class variance, while others generate high-variance faces with low identity consistency. 
In this paper, we propose a Triple Condition Diffusion Model (TCDiff) to improve face style transfer from real to synthetic faces through 2D and 3D facial constraints, enhancing face identity consistency while keeping the necessary high intra-class variance.
Face recognition experiments using 1k, 2k, and 5k classes of our new dataset for training outperform state-of-the-art synthetic datasets in real face benchmarks such as LFW, CFP-FP, AgeDB, and BUPT. Our source code is available at: \textcolor{cyan}{\texttt{\url{https://github.com/BOVIFOCR/tcdiff}}}.
\end{abstract}

\IEEEpeerreviewmaketitle

\section{Introduction}

In recent years, the availability of large face recognition datasets containing thousands of real faces, such as CASIA-WebFace \cite{yi2014learning}, VGGFace2 \cite{vggface2_Parkhi15}, MS1MV3 \cite{Deng_2019_ICCV}, WebFace260M \cite{zhu2021webface260m}, and Glint360K \cite{glint360k_an2022pfc}, has contributed to remarkable advancements in face recognition across various challenging domains, including pose, age, occlusions, and noise. 
With such data, deep neural networks trained with sophisticated angular margin loss functions, such as SphereFace \cite{Liu2017}, CosFace~\cite{Wang2018}, ArcFace~\cite{Deng2019}, CurricularFace~\cite{Huang2020}, MagFace~\cite{Meng2021} and AdaFace~\cite{Kim2022}, have achieved impressive performances on different benchmarks.

However, datasets of this nature present critical ethical, annotation, and bias problems~\cite{bae2023Digiface1m}. 
Furthermore, the long-tailed distribution of samples in many datasets poses additional challenges, necessitating careful network architecture and loss function design to ensure the robustness of model generalization. 
These challenges also make it difficult to explore facial attribute influences like expression, pose, and illumination. In contrast, learning-based face recognition models encode facial images into fixed-dimensional embedding vectors, enabling various tasks like identification and verification.
While publicly available datasets have driven recent progress, they come with associated problems.
Synthetic datasets offer a potential solution, providing privacy benefits, virtually unlimited data generation, and control over demographic characteristics. 
This contrasts with real-world datasets, which are constrained by privacy regulations and representational biases.

Due to such advantages, synthetic faces in face recognition have attracted attention for their potential to mitigate privacy concerns and long-standing dataset biases, such as long-tail distributions and demographic imbalances. Recently, face recognition competitions using synthetic faces have been held, such as FRCSyn \cite{Melzi_2024_WACV} \cite{frcsyn_cvpr2024} and SDFR \cite{shahreza2024sdfr}, showing the increasing interest of the research community on this topic.

\begin{figure}[!t]
\centering
\includegraphics[width=3.5in]{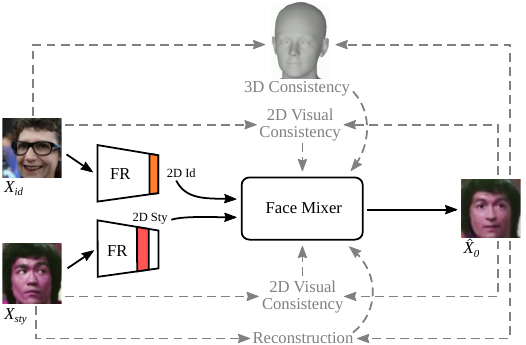}
\vspace*{-7mm}
\caption{Overview of the proposed synthetic face mixer TCDiff with 2D and 3D consistency constraints (gray elements). Intermediate style features are extracted from a style image $X_{sty}$ and applied to a synthetic identity image $X_{id}$, generating a new stylized sample $\hat{X}_{0}$.}
\label{fig_face_mixer}
\end{figure}

Generating synthetic faces and manipulating attributes such as pose, expression, age, noise, and occlusion with high visual fidelity and identity consistency is a challenging task due to the ill-posed nature of representing 3D objects in 2D planes. Therefore, 3D facial constraints might improve model learning and reduce facial inconsistencies.

In this paper, we propose a Triple Condition Diffusion Model (TCDiff) to stylize a synthetic identity face with real style attributes from real faces, such as pose, expression, age, noise, occlusion, shadow, hair, etc., with 2D and 3D consistency constraints, aiming to enhance intra-class identity consistency. 
Fig.~\ref{fig_face_mixer} presents an overview of our method and the constraints computed with identity image ($X_{id}$), style image ($X_{sty}$), and stylized image ($\hat{X}_{0}$). 
Our experimental results show that enhancing intra-class identity consistency improves synthetic dataset quality when training face recognition models with few classes.

\section{Related Work}

Face synthesis has emerged as a prominent area of research, driven by advancements in deep generative models like GANs \cite{8578190, 9156396} and Diffusion Models \cite{NEURIPS2020_4c5bcfec}. Such methods excel at generating high-quality facial images with unique identities. However, they often lack the intra-class variance necessary to train powerful face recognition models. In this regard, recent approaches explore such variance in real-face datasets, mixing synthetic and real images to generate multiple samples from the same synthetic subject.

SynFace \cite{9710281} proposes a Mixup Face Generator designed to create synthetic face images with different identities. To mitigate the domain gap between the synthetic and real face data, the method incorporates a Domain Mixup module regularized by an angular margin loss. In contrast, SFace \cite{10007961} trains a StyleGAN2-ADA \cite{NEURIPS2020_8d30aa96} using identity labels as conditional constraints. 
These constraints are similarly regularized by an angular margin loss.

DigiFace-1M~\cite{bae2023Digiface1m} employs the 3DMM-based model FaceSynthetics~\cite{wood2021fake} to generate multiple synthetic faces, varying their expression and pose parameters. The rendering pipeline from FaceSynthetics further enhances flexibility by allowing modifications in the background and illumination settings in the images. After generating original images, data augmentation techniques including flipping, cropping, adding noise, blurring, and warping are applied to improve the face recognition performance. Despite the 3D consistency in images, this dataset has limitations in appearance due to the intrinsic synthetic texture of faces.

GANDiffFace \cite{10350589} combines the strengths of GANs and Diffusion Models to produce realistic faces while incorporating intra-class variance. Initially, synthetic faces are generated using StyleGAN3 \cite{alaluf2022times} trained on the FFHQ dataset \cite{Karras2018ASG}, and grouped based on by extracted facial attributes such as pose, expression, illumination, gender, and race. Support Vector Machine (SVM) classifiers are trained to distinguish each group. The normal vectors concerning the resulting separation hyperplanes are used as directions to edit facial attributes of faces in latent space. Despite achieving realistic appearances for the generated synthetic faces, there remains a gap in the evaluation performance using real face testing benchmarks of face recognition models, trained on their synthetic dataset.

IDiffFace \cite{Boutros2023IDiffFace} uses a Diffusion Model trained on the FFHQ \cite{Karras2018ASG} real face dataset to create new synthetic faces. These faces are generated using a U-Net-based architecture enriched with residual and attention blocks to encourage the model to improve the intra-class variance generation ability. To prevent overfitting and ensure diverse outputs, they also propose a Contextual Partial Dropout (CPD) technique.

DCFace \cite{10204758} introduces an approach to minimize the distribution gap between synthetic and real face datasets by employing a diffusion model. This model integrates visual constraints to transfer the stylistic characteristics of real faces onto synthetic faces, thereby enhancing the intra-class variance. Initially trained on the CASIA-WebFace \cite{yi2014learning} real face dataset, the model utilizes statistics derived from intermediate features of images, assuming these contain style information to be transferred to any other identity. Despite these efforts, some artifacts and identity inconsistencies persist in the synthetic output.

\begin{figure}[!t]
\centering
\includegraphics[width=3.5in]{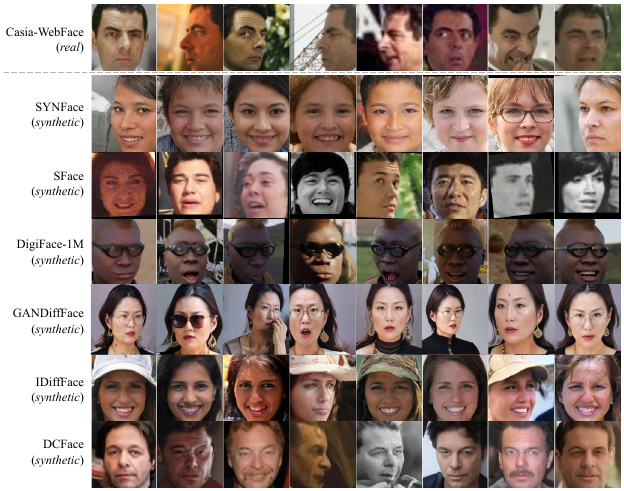}
\vspace*{-7mm}
\caption{Face samples of real (first row) and synthetic datasets. SYNFace has low identity consistency and low intra-class variance; SFace has low identity consistency; DigiFace-1M, GANDiffFace, and IDiffFace have high identity consistency but low intra-class variance; DCFace has low identity consistency and high intra-class variance.}
\label{fig_samples_synthetic_faces}
\end{figure}

Fig.~\ref{fig_samples_synthetic_faces} illustrates some samples generated by the aforementioned methods. Each row corresponds to samples of distinct synthetic subjects, while columns contain different samples of the same subject. While the limited number of images may not be sufficient to provide an accurate visual analysis, they offer some initial intuitions about their characteristics, such as intra-class variance and identity consistency.

\section{Face Style Transfer}

Image style transfer is a technique that generates novel images by merging the content of one image with the stylistic elements of another ~\cite{gatys_stytransf_2016}. This process leverages Convolutional Neural Networks (CNNs) pre-trained on image classification tasks to extract hierarchical intermediate feature representations. The content of an image is captured by the higher layers of the network, which encode the image's structure and objects, while the style is represented by the correlations between feature maps in the lower layers, known as Gram matrices. To achieve style transfer, the technique minimizes a loss function that combines the content loss, which measures the difference between content representations of the original and generated images, and style loss, which quantifies the difference between style representations of the original and generated images. By iteratively adjusting a white noise image based in this combined loss function, the network gradually synthesizes the final output that maintains the content of one image while adopting the style of another.

In the face recognition field, a person's identity can be expressed mainly by facial parts such as eyes, nose, lips, eyebrows, etc., and their spatial position in the face. Meanwhile, style is related to facial pose, expression, age, noise, occlusion, color, etc \cite{suwala_idaware_2024}. Achieving a perfect disentanglement of identity and style representation remains a significant challenge in deep learning. Existing style transfer methods aim to manage this tradeoff depending on the final goal \cite{chen_filterbank_2021}.

\section{Proposed Approach}

For face style extraction, we adopt the proposed model $E_{sty}$ of DCFace \cite{10204758}, which uses intermediate feature maps $I_{sty} \in \mathbb{R}^{C \times H \times W}$ extracted with a pre-trained and fixed weights face recognition model $F_s$ from a given input face image $X_{sty}$, where $C$, $H$ and $W$ are the number of channels, height, and width of the feature maps, respectively. Each feature map is divided into a grid $k \times k$ and each element $I_{sty}^{k_i} \in \mathbb{R}^{C \times \frac{H}{k} \times \frac{W}{k}}$ is mapped on the mean and variance of $I_{sty}^{k_i}$ as

\begin{equation}
    \hat{I}^{k_i} = \text{BN}(\text{Conv}(\text{ReLU}(\text{Dropout}(I_{sty}^{k_i})))),
\end{equation}
\begin{equation}
    \mu_{sty}^{k_i} = \text{SpatialMean}(\hat{I}^{k_i}), \;\;\;  \sigma_{sty}^{k_i} = \text{SpatialStd}(\hat{I}^{k_i}),
\end{equation}
\begin{equation}
    s^{k_i} = \text{LN}((W_1 \odot \mu_{sty}^{k_i} + W_2 \odot \sigma_{sty}^{k_i} ) + P_{emb}),
\end{equation}
\begin{equation}
    E_{sty}(X_{sty}) := s = [s^{1}, s^{2}, s^{k_i}, ..., s^{k \times k}, {s}'], 
\end{equation}

\noindent
where ${s}'$ corresponds to $\hat{I}_{sty}^{k_i}$ being a global feature, where $k = 1$. \ADDED{ $P_{emb} \in \mathbb{R}^{50 \times C}$ is a learned position embedding \cite{pmlr-v70-gehring17a} that makes the model learn to extract patches styles according to their locations in $X_{sty}$ }. BN and LN are BatchNorm \cite{10.5555/3045118.3045167} and LayerNorm \cite{ba2016layernormalization} operations.

\begin{figure}[!t]
\centering
\includegraphics[width=3.5in]{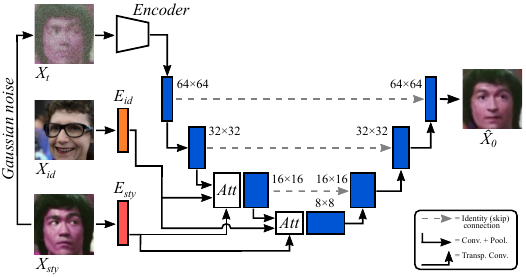}
\vspace*{-7mm}
\caption{\ADDED{Simplified illustration of internal architecture of the U-Net DDPM face mixer $\epsilon_{\theta}$. Intermediate feature maps (blue boxes) are extracted from $X_t$, a noisy version of $X_{sty}$, by encoders and copied to their corresponding upscale layer. Identity features $E_{id}$, extracted from $X_{id}$, and style features $E_{sty}$, extracted from $X_{sty}$, are fed to attention modules together with intermediate feature maps, allowing the model learn how to denoise $X_t$ with features from $X_{id}$ and $X_{sty}$.}}
\label{face_mixer_unet}
\end{figure}

The face style embedding \ADDED{$E_{sty}$} extracted from $X_{sty}$ is then applied to an identity image $X_{id}$ using a U-Net denoising diffusion probabilistic model (DDPM) \cite{NEURIPS2020_4c5bcfec} face mixer $\epsilon_{\theta}(X_t, t, E_{id}(X_{id}), E_{sty}(X_{sty}))$, \ADDED{whose architecture is illustrated in Fig. \ref{face_mixer_unet}}. $X_t$ is a noisy version of $X_{sty}$ at time-step $t$ and $E_{id}$ is a face recognition model ResNet50 \cite{he2016deep} responsible for extracting a discriminant identity embedding.

Given an identity image $X_{id}$ and a style image $X_{sty}$, a new stylized image $\hat{X}_{0}$ is obtained as

\begin{equation}
    \hat{X}_{0} = (X_t - \sqrt{1 - \bar{\alpha}} \epsilon_{\theta}(X_t, t, X_{id}, X_{sty})) / \sqrt{\bar{\alpha}_t}.
\end{equation}

\noindent
where $\bar{\alpha}_t$ is a pre-set variance scheduling scalar \cite{NEURIPS2020_4c5bcfec}.

To train the face mixer $\epsilon_{\theta}$, we first employ the mean squared error (MSE) loss $L_{MSE}$ between style image $X_{sty}$ and stylized image $\hat{X}_{0}$

\begin{equation}
    L_{MSE} = \frac{1}{MN} \sum_{i=1}^{M} \sum_{j=1}^{N} \left( X_{sty}(i,j) - \hat{X_0}(i,j) \right)^2
\end{equation}

\noindent
to enforce the model to preserve relevant style features of $X_{sty}$ in $\hat{X_0}$. Additionally, to balance identity and style features of $X_{id}$ and $X_{sty}$ in $\hat{X}_{0}$, we also employ an identity loss $L_{ID}$ through cosine similarity (CS)

\begin{equation}
\begin{split}
    L_{ID} = &- \gamma_t \text{CS}(F(X_{id}), F(\hat{X}_0)) \\
             &- (1-\gamma_t)\text{CS}(F(X_{sty}), F(\hat{X}_0))
\end{split}
\end{equation}

\noindent
where $\gamma_t \in \mathbb{R} ~ | ~ 0 \leq \gamma_t \leq 1$.

To enhance intra-class identity consistency when stylizing synthetic faces, we also propose to add a 3D facial shape loss $L_{3\text{D}}$

\begin{equation}
    L_{3\text{D}} = \sqrt{\sum (x_{id}^{3\text{D}} - \hat{x}_{0}^{3\text{D}})^2}
\end{equation}

\noindent
which computes the Euclidean Distance between \ADDED{the} 3DMM \cite{Blanz_3Dmm_1999} shape feature vectors $x_{id}^{3\text{D}}$ and $\hat{x}_{0}^{3\text{D}}$, obtained from $X_{id}$ and $\hat{X}_0$. Due to the lack of large datasets containing both 2D and 3D scanned representations of real facial, we obtained 3D Morphable Model (3DMM) coefficients during training using the 3D face reconstruction model MICA \cite{MICA:ECCV2022}.

\begin{figure*}[!b]
\centering
\includegraphics[width=\textwidth]{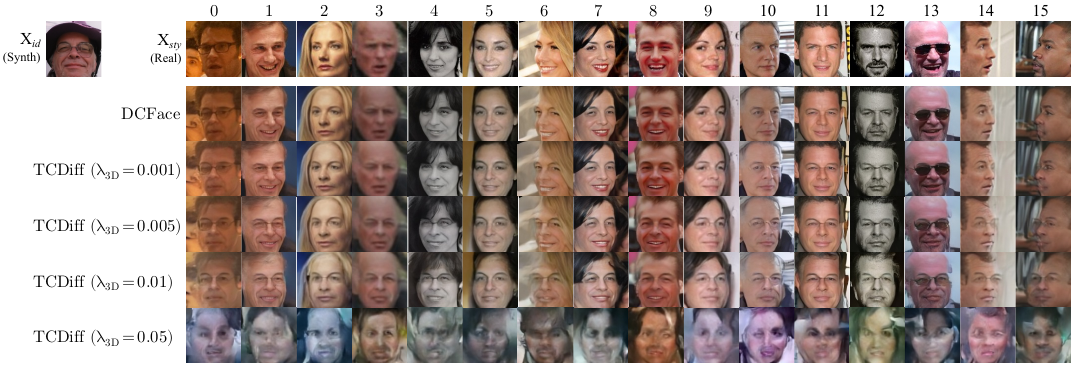}
\vspace*{-7mm}
\caption{Stylized synthetic face samples generated with DCFace \cite{10204758} and our proposed TCDiff face mixer. The first row shows the original synthetic $X_{id}$ face and $16$ real style faces $X_{sty}$ used to create new samples. In our experiments, $\lambda_{3\text{D}}=0.001$ is the best value to balance intra-class identity consistency and variance.}
\label{fig_samples_id_sty}
\end{figure*}

\begin{figure*}[!b]
\centering
\includegraphics[width=\textwidth]{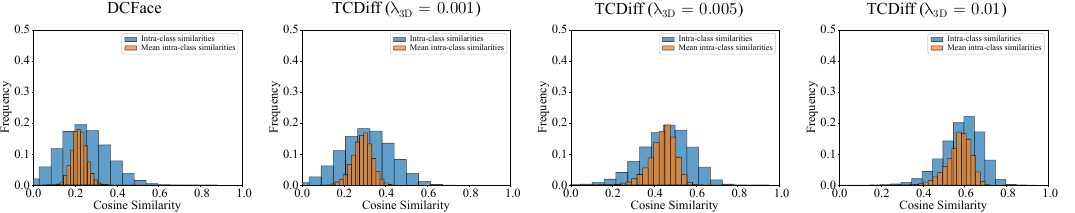}
\vspace*{-7mm}
\caption{Intra-class cosine similarities of synthetic datasets generated by DCFace \cite{10204758} and our proposed model TCDiff with different values of $\lambda_{3\text{D}}$, computed with ResNet100/Arcface \cite{Deng2019} trained on MS1MV3 \cite{Deng_2019_ICCV}. The higher the $\lambda_{3\text{D}}$ value, the higher the intra-class identity consistency.}
\label{fig_intraclass_distances}
\end{figure*}

The shape of a face in a 3DMM representation is described by the positions of a set of 3D vertices $S = (x_1, y_1, z_1, x_2, y_2, z_2, \dots, x_n, y_n, z_n)^T \in \mathbb{R}^{3n}$. These vertices form a mesh that captures the geometric structure of the face. Mathematically, the shape vector $S$ can be expressed as a linear combination of a mean shape $\bar{S}$ and a set of shape basis vectors $S_i$:

\begin{equation}
    S = \bar{S} + \sum_{i=1}^{n} \alpha_i S_i,
\end{equation}

\noindent
where $\alpha_i$ are the shape coefficients that determine how much each basis vector $S_i$ contributes to the final shape. The shape vector $S$ consists of the 3D coordinates of all vertices in the mesh.

Similarly, the face texture is described by a vector $T = (R_1, G_1, B_1, R_2, G_2, B_2, \dots, R_n, G_n, B_n)^T \in \mathbb{R}^{3n}$, which is a linear combination of a mean texture $\bar{T}$ and a set of texture basis vectors $T_i$:

\begin{equation}
    T = \bar{T} + \sum_{i=1}^{m} \beta_i T_i,
\end{equation}

\noindent
where $\beta_i$ are the texture coefficients that control the contribution of each texture basis vector $T_i$. The texture vector $T$ consists of the RGB color values for each vertex in the mesh.

The basis vectors $S_i$ and $T_i$, for shape and texture, are derived from a Principal Component Analysis (PCA) over a set of real 3D face scans, which allows representing new faces within the training set variance.

To obtain the 3DMM facial shape coefficients $x_{id}^{3\text{D}}$ and $\hat{x}_{0}^{3\text{D}}$ from identity face $X_{id}$ and stylized face $\hat{X_{0}}$, the MICA \cite{MICA:ECCV2022} method uses the face embedding produced by a state-of-the-art 2D face recognition network \cite{Deng_subarcface_2020} as input to a small mapping network $z = M(ArcFace(I)) \in \mathbb{R}^{300}$. \ADDED{Therefore, $x_{id}^{3\text{D}} = M(X_{id})$ and $\hat{x}_{0}^{3\text{D}} = M(\hat{X_{0}})$}.

Finally, our total loss function $L_{T}$ is defined as

\begin{equation}
    L_{T} = L_{MSE} + \lambda_{id} L_{ID} + \lambda_{3\text{D}} L_{3\text{D}},
\end{equation}

\noindent
where $\lambda_{id}$ and $\lambda_{3\text{D}}$ are scaling parameters to balance the importance of 2D and 3D facial identity constraints.

\section{Experiments}

This section presents our experimental setup, datasets, obtained results, and qualitative analysis. To fairly evaluate the robustness of our proposed TCDiff face mixer, we adopted the same protocol of DCFace \cite{10204758} by using the real faces dataset CASIA-WebFace \cite{yi2014learning} as the training set and a grid 5 $\times$ 5 for style feature extraction. Our model was trained for 10 epochs with batch=$16$ using AdamW Optimizer \cite{Kingma2014AdamAM} with the learning rate of $1\mathrm{e}{-4}$ on one GPU NVIDIA GeForce RTX 3090. We set $\lambda_\text{ID}=0.05$ and varied $\lambda_{3\text{D}}=\{0.001, 0.005, 0.01, 0.05\}$ to analyse the impact of $3\text{D}$ consistency constraints.

After training TCDiff, we selected the same 10k distinct synthetic identities of DCFace \cite{10204758}, which were generated using the publicly released unconditional DDPM \cite{ho_ddpm_2020} trained on FFHQ \cite{karras_ffhq_2021}. Each synthetic identity was stylized with 50 randomly chosen real faces of CASIA-WebFace \cite{yi2014learning}, resulting in a new synthetic dataset of 500k images. Fig. \ref{fig_samples_id_sty} shows 1 synthetic face image $X_{id}$, 16 real faces $X_{sty}$ from CASIA-WebFace, and their corresponding 16 new samples of $X_{id}$.

We choose ResNet50~\cite{he2016deep} backbone and ArcFace~\cite{Deng2019} loss as Face Recognition (FR) model to evaluate the quality of the proposed synthetic dataset in cross-dataset scenarios using seven different datasets in face verification (1:1) task: LFW \cite{LFWTech} CFP-FF \cite{cfp_paper}, CPLFW \cite{CPLFWTech}, CFP-FP \cite{cfp_paper}, AgeDB \cite{moschoglou2017agedb}, CALFW \cite{zheng_calfw_2017}, and BUPT-CBFace \cite{zhang2020class}.
These datasets are commonly applied in FR to validate or test models. Each dataset contains a verification protocol consisting of face pairs labeled as \textit{genuine} (same person) or \textit{impostor} (different person).

LFW (6k pairs) and CFP-FF (7k pairs) protocols are mainly focused on frontal face verification, representing controlled scenarios. Otherwise, CPLFW (6k pairs) and CFP-FP (7k pairs) contain faces with more varied poses to simulate in-the-wild scenarios. AgeDB (7k pairs) and CALFW (6k pairs) focus on comparing faces with large age differences, while BUPT-CBFace (8k pairs) contains the same number of pairs of 4 distinct ethnic groups: Asian, Caucasian, African, and Indian. All protocols were split into 10-fold containing 50$\%$ of genuine and 50$\%$ of impostor pairs. Following the cross-validation method, we use $9$ folds to select the best threshold and 1 for the final test.

\section{Discussion}

In this section, we present a qualitative and quantitative analysis of the results obtained with the synthetic datasets generated with DCFace \cite{10204758} and our proposed model TCDiff for the face recognition task.

Even with few samples, we can visually observe in Fig.~\ref{fig_samples_id_sty} that synthetic faces stylized by DCFace~\cite{10204758} have low intra-class consistency, compared to its corresponding original synthetic identity $X_{id}$. For instance, the stylized male faces 0, 8, and 10 seem to belong to distinct identities. 
The same happens with the stylized female faces 2 and $9$. 
Otherwise, our face mixer $\epsilon_{\theta}$ tends to preserve identity regions, such as eyes, nose, and mouth of original $X_{id}$ in stylized faces to enhance the intra-class consistency imposed by $L_{3\text{D}}$ when $\lambda_{3\text{D}}$ varies from $0.001$ to $0.01$.
Stylized faces are completely degraded when $\lambda_{3\text{D}} = 0.05$ and this setting was ignored in our experiments.

\begin{table*}[!t]
\caption{Face verification results \ADDED{(\%)} of \ADDED{ResNet50} on LFW \cite{LFWTech}, CFP-FF \cite{cfp_paper}, CPLFW \cite{CPLFWTech}, CFP-FP \cite{cfp_paper}, AgeDB \cite{moschoglou2017agedb}, CALFW \cite{zheng_calfw_2017} and BUPT-CBFace \cite{zhang2020class} when training on different synthetic face datasets generated with DCFace \cite{10204758} and our proposed model TCDiff. The best results are in bold.}
\label{tab-fr-results}
\vspace*{-3mm}
\resizebox{\textwidth}{!}{%
\renewcommand{\arraystretch}{0.9} %
\begin{tabular}{@{}clllllllllccc@{}}
\toprule
\#Train Subj. & Train Dataset & LFW    & CFP-FF  & CPLFW    & CFP-FP   & AgeDB    & CALFW    & BUPT     & &  Avg     & Std               & Avg-Std \\
\midrule
 & DCFace        & 88.30 & 88.70 &  61.00 &  62.40 &  70.10 &  76.60 &  75.30 & &  74.63 &  \textbf{11.14} &  63.49 \\
 \multirow{-0.5}{*}{1k}
 & TCDiff $(\lambda_{3\text{D}}=0.001)$ & 90.50 & 91.30 & 62.50 & 64.20 & 72.40 & 80.20 & 80.30 & & 77.34 & 11.56 & 65.78 \\ 
 & \ADDED{TCDiff $(\lambda_{3\text{D}}=0.005)$} & \ADDED{\textbf{90.55}} & \ADDED{\textbf{91.87}} & \ADDED{\textbf{63.83}} & \ADDED{\textbf{64.42}} & \ADDED{\textbf{73.10}} & \ADDED{\textbf{81.40}} & \ADDED{\textbf{80.67}} & & \ADDED{\textbf{77.98}} & \ADDED{11.39} & \ADDED{\textbf{66.59}} \\
 & \ADDED{TCDiff $(\lambda_{3\text{D}}=0.01)$} & \ADDED{88.05} & \ADDED{89.47} & \ADDED{58.71} & \ADDED{59.42} & \ADDED{67.38} & \ADDED{78.61} & \ADDED{78.05} & & \ADDED{74.24} & \ADDED{12.68} & \ADDED{61.56} \\

\midrule
 & DCFace &  92.80 &  94.07 &  67.48 &  71.53 &  77.07 &  83.55 &  82.26 & &  81.25 & \textbf{10.05} & 71.20 \\
 \multirow{-0.5}{*}{2k}
 & TCDiff $(\lambda_{3\text{D}}=0.001)$ & \textbf{94.22} & \textbf{95.07} & \textbf{68.18} & \textbf{72.03} & \textbf{79.13} & \textbf{85.03} & \textbf{84.14} & & \textbf{82.54} & 10.25 & \textbf{72.29} \\
 & \ADDED{TCDiff $(\lambda_{3\text{D}}=0.005)$} & \ADDED{93.15} & \ADDED{94.22} & \ADDED{67.05} & \ADDED{67.08} & \ADDED{76.70} & \ADDED{84.26} & \ADDED{83.41} & & \ADDED{80.84} & \ADDED{11.15} & \ADDED{69.69} \\
 & \ADDED{TCDiff $(\lambda_{3\text{D}}=0.01)$} & \ADDED{90.70} & \ADDED{92.11} & \ADDED{61.84} & \ADDED{61.08} & \ADDED{71.63} & \ADDED{81.81} & \ADDED{81.25} & & \ADDED{77.20} & \ADDED{12.71} & \ADDED{64.49} \\

\midrule
 & DCFace & 96.40 & 96.70 & 72.00 & 79.10 & 82.60 & 87.30 & 86.40 & & 85.79 & \:\;8.94 & 76.85 \\
\multirow{-0.5}{*}{5k}
 & TCDiff $(\lambda_{3\text{D}}=0.001)$ & \textbf{96.87} & \textbf{97.26} & \textbf{73.92} & \textbf{79.86} & \textbf{83.68} & \textbf{87.92} & \textbf{89.18} & &  \textbf{86.96} & \textbf{\:\;8.58} & \textbf{78.38} \\
 & \ADDED{TCDiff $(\lambda_{3\text{D}}=0.005)$} & \ADDED{94.30} & \ADDED{95.11} & \ADDED{68.40} & \ADDED{69.60} & \ADDED{78.60} & \ADDED{86.20} & \ADDED{84.60} & &  \ADDED{82.40} & \ADDED{10.78} & \ADDED{71.63} \\
 & \ADDED{TCDiff $(\lambda_{3\text{D}}=0.01)$} & \ADDED{92.31} & \ADDED{93.74} & \ADDED{64.75} & \ADDED{63.81} & \ADDED{74.96} & \ADDED{84.01} & \ADDED{83.13} & &  \ADDED{79.53} & \ADDED{12.14} & \ADDED{67.39} \\

\midrule
 & DCFace & \textbf{98.02} & 97.73 & \textbf{79.62} & \textbf{85.01} & \textbf{88.82} & \textbf{90.48} & \textbf{91.35} & & \textbf{90.15} & \textbf{\:\;6.58} & \textbf{83.56} \\
\multirow{-0.5}{*}{10k}
 & TCDiff $(\lambda_{3\text{D}}=0.001)$ & 97.77 & \textbf{98.04} & 78.13 & 82.21 & 86.35 & 90.33 & 90.73 & & 89.08 & \:\;7.47 & 81.61 \\
 & \ADDED{TCDiff $(\lambda_{3\text{D}}=0.005)$} & \ADDED{95.98} & \ADDED{96.69} & \ADDED{70.95} & \ADDED{71.83} & \ADDED{81.13} & \ADDED{87.72} & \ADDED{87.66} & & \ADDED{84.57} & \ADDED{10.46} & \ADDED{74.11} \\
 & \ADDED{TCDiff $(\lambda_{3\text{D}}=0.01)$} & \ADDED{93.51} & \ADDED{95.10} & \ADDED{64.95} & \ADDED{64.64} & \ADDED{74.68} & \ADDED{85.01} & \ADDED{84.20} & & \ADDED{80.30} & \ADDED{12.54} & \ADDED{67.76} \\
 
\bottomrule
\end{tabular}%
}
\end{table*}

Such a qualitative analysis is quantitatively confirmed in Fig.~\ref{fig_intraclass_distances}, where intra-class cosine similarities are presented.
\ADDED{Blue bars show the distributions of all $\sum_{i=0}^{M} C^{N_{i}}_{2}$ similarities, where $M$ is the number of classes and $N_i$ is the number of samples of the $i$-$th$ class, while orange bars show the distributions of the mean intra-class similarities}.
One can observe higher similarities in faces stylized with our model TCDiff, indicating a higher intra-class consistency \ADDED{due to the shape of eyes, nose, lips, skin color, pose, expression, and facial accessories}. 

\ADDED{The quality of a face dataset for FR task is assessed not just by the images themselves, but by the performances of FR models trained on it. Results in Table~\ref{tab-fr-results} show that enhancing intra-class consistency improves synthetic datasets quality when training with 1k, 2k, and 5k classes. We hypothesize that such an identity consistency improves inter-class separability when the number of distinct identities are low.}

However, this improvement is surpassed when training with 10k classes, indicating that the inter-class variability is also an important property of synthetic datasets for face recognition.
\ADDED{ResNet50 performed slightly better on dataset CFP-FF \cite{cfp_paper} when training with stylized images by our model TCDiff $(\lambda_{3\text{D}}=0.001)$, due to the greater existence of frontal faces in such a dataset.}

\section{Conclusion and Future Work}

In this work we propose TCDiff, a face style transfer trained with 2D and 3D facial constraints, aiming to improve the quality of synthetic datasets for face recognition. By increasing the importance of 3D constraints, our model can preserve identity features of the input synthetic face to be stylized, which enhances intra-class identity consistency.

This behavior contributes to increasing the quality of small synthetic datasets and might be explored in the future for more classes, as the interest in this field is growing due to the advantages of synthetic data. As a future step, facial expression and pose constraints might be added to the face styler model, aiming to balance better the importance of identity and style features in newly generated samples.

\section*{Acknowledgment}

This work was supported by
\iffinal a tripartite-contract, i.e., unico - idTech, UFPR (Federal University of Paraná), and FUNPAR (Fundação da Universidade Federal do Paraná).  
We thank the Federal Institute of Mato Grosso (IFMT), Pontes e Lacerda, for supporting Bernardo Biesseck, and also thank the National Council for Scientific and Technological Development (CNPq) (\# 315409/2023-1) for supporting Prof. David Menotti.
\else
ANONYMOUS.
\fi

\bibliographystyle{IEEEtran}
\bibliography{example}

\end{document}